\documentclass[conference, 10pt]{IEEEtran}
\IEEEoverridecommandlockouts

\usepackage{cite}
\usepackage{amsmath,amssymb,amsfonts}
\usepackage{graphicx}
\usepackage{textcomp}
\usepackage{xcolor}
\usepackage{url}
\usepackage{placeins}
\usepackage{float}
\usepackage{tabularx,colortbl}
\usepackage{ifthen}
\usepackage{caption,subcaption}
\usepackage{booktabs}
\renewcommand{\thetable}{\Roman{table}}
\let\labelindent\relax
\usepackage{enumitem}
\usepackage{stfloats}
\usepackage{wrapfig}
\usepackage{verbatim}
\usepackage{tabularx}
\usepackage{booktabs}
\usepackage{multirow}
\usepackage{caption}
\usepackage{flushend}
\usepackage{algorithm}
\usepackage{algpseudocode}
\usepackage{hyperref}  
\usepackage{cleveref}  

\def\BibTeX{{\rm B\kern-.05em{\sc i\kern-.025em b}\kern-.08em
    T\kern-.1667em\lower.7ex\hbox{E}\kern-.125emX}}

\begin{document}
\newcolumntype{P}[1]{>{\centering\arraybackslash}p{#1}}
\newcolumntype{M}[1]{>{\centering\arraybackslash}m{#1}}
\setlength{\textfloatsep}{10pt plus 1.0pt minus 3.0pt}
\setlength{\dbltextfloatsep}{10pt plus 1.0pt minus 3.0pt}
\setlength{\floatsep}{10pt plus 1.0pt minus 3.0pt}
\setlength{\dblfloatsep}{10pt plus 1.0pt minus 3.0pt}
\setlength{\intextsep}{10pt plus 1.0pt minus 3.0pt}


\title{\LARGE \bf Sense Less, Generate More: Pre-training LiDAR Perception with Masked Autoencoders for Ultra-Efficient 3D Sensing}
\author{Sina Tayebati, Theja Tulabandhula, and Amit R. Trivedi
\thanks{ Authors are with the University of Illinois Chicago (UIC), Chicago, IL, Email: {\tt\small amitrt@uic.edu}}
}
\maketitle

\begin{abstract}
In this work, we propose a disruptively frugal LiDAR perception dataflow that \textit{generates} rather than \textit{senses} parts of the environment that are either predictable based on the extensive training of the environment or have limited consequence to the overall prediction accuracy. Therefore, the proposed methodology \textit{trades off sensing energy with training data} for low-power robotics and autonomous navigation to operate frugally with sensors, extending their lifetime on a single battery charge. Our proposed generative pre-training strategy for this purpose, called as radially masked autoencoding (R-MAE), can also be readily implemented in a typical LiDAR system by selectively activating and controlling the laser power for randomly generated angular regions during on-field operations. Our extensive evaluations show that pre-training with R-MAE enables focusing on the radial segments of the data, thereby capturing spatial relationships and distances between objects more effectively than conventional procedures. Therefore, the proposed methodology not only reduces sensing energy but also improves prediction accuracy. For example, our extensive evaluations on Waymo, nuScenes, and KITTI datasets show that the approach achieves over a 5\% average precision improvement in detection tasks across datasets and over a 4\% accuracy improvement in transferring domains from Waymo and nuScenes to KITTI. In 3D object detection, it enhances small object detection by up to 4.37\% in AP at moderate difficulty levels in the KITTI dataset. Even with 90\% radial masking, it surpasses baseline models by up to 5.59\% in mAP/mAPH across all object classes in the Waymo dataset. Additionally, our method achieves up to 3.17\% and 2.31\% improvements in mAP and NDS, respectively, on the nuScenes dataset, demonstrating its effectiveness with both single and fused LiDAR-camera modalities. Codes are publicly available at \textcolor{magenta}{\emph{\url{https://github.com/sinatayebati/Radial_MAE}}}.
\end{abstract}


\section{Introduction}
Multispectral sensors such as LiDARs (Light Detection and Ranging) excel in depth perception and object detection across various lighting conditions, including complete darkness and bright sunlight. Unlike cameras, LiDAR outputs are not affected by optical illusions or ambient light variations, making them more reliable for accurate environmental mapping. Thus, LiDARs have become essential for autonomous navigation and robotics \cite{zou2021comparative}. However, due to their \textit{active sensing}--where they radiate the environment and measure the reflections--LiDARs are also much more energy-intensive than cameras. For instance, among state-of-the-art LiDAR systems, Velodyne's Velarray H800 LiDAR sensor consumes approximately 13 watts \cite{schulte2022benchmarking}, Luminar's LiDAR up to 25 watts \cite{rablau2019lidar}, InnovizPro's solid-state LiDAR 10-12 watts \cite{schulte2022benchmarking}, LeddarTech Leddar Pixell around 15 watts \cite{deziel2021pixset}, and Quanergy's M8 LiDAR up to 12 watts \cite{mittet2016experimental}. Comparatively, modern digital cameras require only about 1-2 watts \cite{sahin2019long}, making LiDAR-based autonomy prohibitively more energy-expensive for low power robotics applications that require prolonged operational periods with minimal battery resources.

\begin{figure*}[t!]
    \centering
    \includegraphics[width=0.28\textwidth]{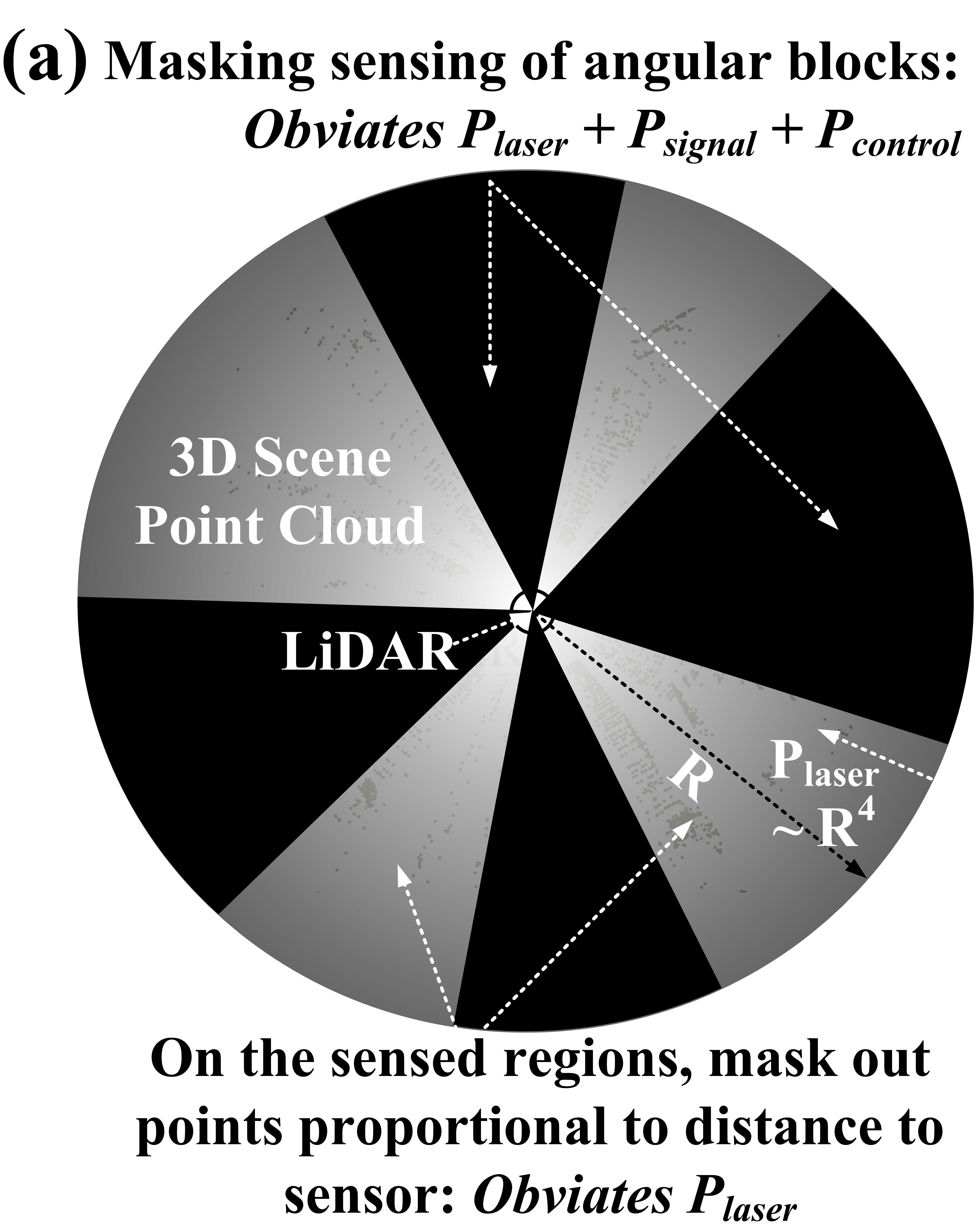}
    \includegraphics[width=0.7\textwidth]{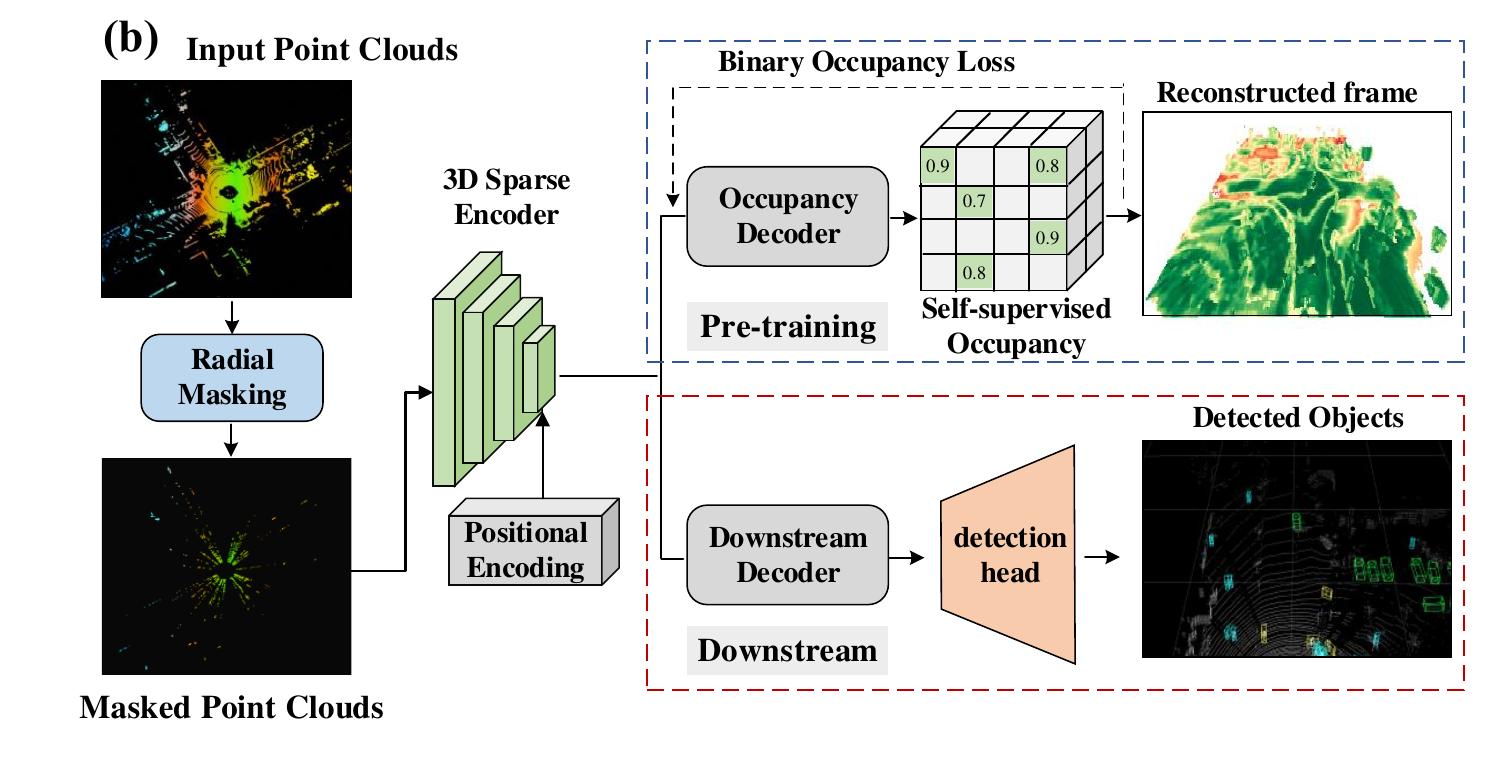}
    \caption{\textbf{(a) Radially masked autoencoding (R-MAE) strategy:} Angular regions are completely masked out as shown in the black by turning off laser emissions. Even on the sensed regions, points are probabilistically dropped-off proportional to their distance $R$. Notably, $P_\text{laser} \sim R^4$, therefore the pretraining encourages models to predict accurately with a low power laser. \textbf{(b) R-MAE processing flow:} The input point cloud is voxelized and radially masked based on voxel distance from the sensor. A 3D spatially sparse convolutional encoder extracts latent features from unmasked voxels, while a decoder reconstructs the 3D scene by predicting voxel occupancy via binary classification.}
    \label{network}
\end{figure*}

In this work, addressing the energy challenges of LiDAR for low power robotics using generative AI, we propose a LiDAR perception system that \textit{generates} rather than \textit{senses} parts of the environment that are either predictable based on the extensive training of the environment or have limited consequence to the overall prediction accuracy [\autoref{network}]. Thus, by only measuring the environment minimally and using generative models to \textit{fill in the blanks}, LiDAR's energy consumption can be dramatically minimized. While comparable prior works \cite{yang2023gd, min2023occupancy, xu2023mv, krispel2024maeli} to our approach leverage generative AI models to compress point cloud data into lower-dimensional latent spaces to facilitate faster and more efficient downstream processing tasks, they miss upon the important opportunity to minimize the sensor power itself using generative-AI. By rather focusing on minimizing sensing energy than feature dimensions, our approach aligns with a notable trend in semiconductor technology advancements, where computing energy decreases at a much faster rate than sensing energy \cite{ajani2024advancements, li2020energy}. The computing energy of a digital technology is determined by how it represents binary bits `0' and `1,' and with each new generation of transistors and emerging technologies, the energy of binary representations continues to dramatically reduce. Meanwhile, sensing energy, especially for active sensing, is more fundamentally constrained by environmental factors such as atmospheric absorption, signal scattering, and reflection characteristics \cite{bijelic2018benchmark, schulte2022benchmarking}. Therefore, prioritizing sensing power using generative AI could offer significantly greater benefits compared to the current methods.

Leveraging generative models to maximize LiDAR efficiency, we introduce R-MAE (\autoref{network}), a generative pre-training paradigm that employs a masked autoencoder with a novel range-aware radial masking strategy. R-MAE effectively expands visible regions by predicting voxel occupancy in masked areas, utilizing rich feature representations learned by the encoder. This reduces the need for extensive sensing by combining partial observation with a pre-trained generative model to reconstruct the 3D scene. Extensive experiments on Waymo \cite{sun2020scalability}, nuScenes \cite{caesar2020nuscenes}, and KITTI \cite{geiger2012we} demonstrate that R-MAE preserves spatial continuity and encourages viewpoint invariance even with 90\% masking. Training with the masking strategy also allows the model to focus on the radial aspects of the data, thus in capturing the spatial relationships and distances between objects more effectively in a 3D scene. Our approach thereby achieved over 5\% average precision improvement in detection tasks across Waymo and other datasets while achieving over 4\% accuracy improvements in transferring domains from Waymo and nuScenes to KITTI.

\section{Trading off \textit{LiDAR Energy} with \textit{Data} using Generative Pretraining}

A LiDAR system incurs power consumption for laser emission, scanning, signal processing, and data acquisition/control, thus requires \(P_{\text{total}} = P_{\text{laser}} + P_{\text{scan}} + P_{\text{signal}} + P_{\text{control}}\) for its overall operations. The laser emitter's power, \(P_{\text{laser}}\), depends on the energy per pulse, \(E_{\text{pulse}}\), the pulse repetition frequency, \(f_{\text{pulse}}\), and the laser efficiency, \(\eta_{\text{laser}}\), \(P_{\text{laser}} = \frac{E_{\text{pulse}} \times f_{\text{pulse}}}{\eta_{\text{laser}}}\). For mechanical scanning systems, the power consumption, \(P_{\text{scan}}\), depends on the voltage supplied to the motor, \(V_{\text{motor}}\), the current drawn by the motor, \(I_{\text{motor}}\), and the motor efficiency, \(\eta_{\text{motor}}\), \(P_{\text{scan}} = \frac{V_{\text{motor}} \times I_{\text{motor}}}{\eta_{\text{motor}}}\). In MEMS or solid-state LiDAR systems, \(P_{\text{scan}}\) is consumed to actuate MEMS mirrors or phase arrays. The power consumption for signal processing, \(P_{\text{signal}}\), depends on the computational complexity and processing architecture. Control and data acquisition power consumption, \(P_{\text{control}}\), is incurred for data handling, system control, and communication, \(P_{\text{control}} = P_{\text{ADC}} + P_{\text{MCU}}\), where \(P_{\text{ADC}}\) is the power consumption of the analog-to-digital converters, and \(P_{\text{MCU}}\) is the power consumption of the microcontroller unit.

\textit{Importantly}, the above energy components are subjected to fundamental \textit{energy-accuracy-range trade-offs}. For instance, at increasing range \(R\), \(E_{\text{pulse}}\) increases as \(E_{\text{pulse}} = \frac{P_r \cdot (4 \pi R^2)^2 \cdot \tau}{A_r \cdot \rho \cdot \eta}\), where \(A_r\) is the area of the receiver aperture, \(\rho\) is the target reflectivity, \(\eta\) is the system efficiency, and \(\tau\) is the laser pulse width. Since the minimum received signal strength \(P_r\) cannot be arbitrarily small, the necessary transmission energy increases as \(R^4\) with increasing range. Range resolution, \(\Delta R\), in LiDAR refers to its ability to distinguish between two closely spaced objects along the line of sight. \(\Delta R\) is controlled by the pulse width (\(\tau\)), where a shorter pulse width allows for finer range resolution, \(\Delta R = \frac{c \cdot \tau}{2}\), \(c\) is the speed of light. Therefore, achieving higher precision (smaller \(\Delta R\)) requires a higher energy per pulse \(E_{\text{pulse}}\) for a given target reflectivity and system efficiency. Increasing \(E_{\text{pulse}}\) is challenging due to power supply and thermal management constraints and necessitates advanced solutions to prevent overheating and power supply variations due to peak demand \cite{lee2020accuracy}. Likewise, angular precision of LiDAR, \(\Delta \theta\), refers to the accuracy with which the system can measure and distinguish angles between objects. \(\Delta \theta\) is determined by beam divergence and the diameter of the laser aperture \(D\), \(\Delta \theta = \frac{\lambda}{D}\), where \(\lambda\) is the wavelength of the laser. To achieve finer angular precision (smaller \(\Delta \theta\)), a larger aperture diameter \(D\) is required, which in turn increases LiDAR's footprint. Alternatively, a lower \(\lambda\) for finer precision is constrained by issues such as eye safety, atmospheric absorption, due to much higher energy of transmitted waves \cite{raj2020survey}.

Higher range and range resolution also require higher ADC sampling rates and more bits for accurate data representation, resulting in higher power consumption of the ADC (\(P_{\text{ADC}}\)). For accurate signal capture, the ADC sampling rate (\(f_s\)) must exceed the Nyquist rate, which is twice the pulse repetition frequency (\(f_{\text{pulse}}\)): \(f_s \geq 2 f_{\text{pulse}}\). Given that \(f_{\text{pulse}}\) is inversely proportional to the desired range resolution, \(f_{\text{pulse}} \approx \frac{c}{2 \Delta R}\), the sampling rate becomes \(f_s \geq \frac{c}{\Delta R}\). The power consumption of the ADC can therefore be approximated by \(P_{\text{ADC}} = k \cdot \frac{c}{\Delta R} \cdot 2^N\), where \(k\) is a constant dependent on the ADC technology. Thus, as the range resolution (\(\Delta R\)) improves (decreases), the sampling rate (\(f_s\)) increases, leading to higher ADC power consumption. Signal processing power \(P_{\text{signal}}\) also increases with higher range and range/angular resolution due to increasing computational complexity and the data rate. For example, FFT (Fast Fourier Transform) operations have a complexity of \(O(N \log N)\), where \(N\) is the number of samples. With higher range resolution, thus higher \(P_{\text{signal}}\) is incurred.

Due to such intricate interactions among various LiDAR performance metrics, simultaneously achieving high precision, extended range, low footprint, energy efficiency, and cost-effectiveness is challenging for most LiDAR systems. Meanwhile, emerging robotics applications demand both low power for prolonged operational periods as well as high safety standards, necessitating novel solutions that can operate with high range, high-precision LiDAR systems without imposing significant energy constraints. E.g., autonomous drones require efficient power usage for extended flight times while ensuring collision avoidance, agricultural robots need to navigate and perform tasks in vast fields with minimal battery drain, and medical robots must function reliably in sensitive environments without frequent recharging. Addressing these challenges for energy efficient and high performance LiDAR perception for low-power robotics, subsequently, we present a novel generative pretraining that \textit{trades off sensing energy with training data} to maximize LiDAR systems' performance within limited energy operations.  

\section{Radially Masked Autoencoding (R-MAE) of LiDAR Scans}
While random masking has proven effective in pre-training models for various modalities \cite{devlin2018bert, he2022masked, pang2022masked}, its direct application to large-scale LiDAR point clouds is challenging. LiDAR data is inherently irregular and sparse, making conventional block-wise masking less effective and potentially requiring substantial hardware modifications for real-time implementation. To address these issues, we propose Radially Masked Autoencoding (R-MAE). This approach masks random angular portions of a LiDAR scan (\autoref{network}(a)) and leverages an autoencoder to predict the occupancy of these unsensed regions. Pre-training R-MAE on unlabeled point clouds allows it to capture underlying geometric and semantic structures. This pre-trained model is then fine-tuned with detection heads, enhancing downstream accuracy by incorporating inductive biases learned from large-scale data. By generating, rather than sensing, a significant portion of the environment, R-MAE reduces LiDAR scan requirements, minimizing energy consumption in laser emission \(P_{\text{laser}}\), data conversion \(P_{\text{ADC}}\), and signal processing \(P_{\text{signal}}\). Importantly, R-MAE also extracts high-level semantic features without relying on labeled data, improving detection accuracy compared to conventional training. Additionally, R-MAE is readily implementable on modern LiDAR systems with programmable interfaces, enabling selective laser activation during inference. Key components of R-MAE are detailed below: 

\subsection{\textbf{Radial Masking Strategy}} To efficiently process large-scale LiDAR point clouds while mimicking the sensor's radial scanning mechanism, we employ a voxel-based radial masking strategy \cite{yan2018second, zhu2021cylindrical, darabi2023starnet}. The point cloud is initially voxelized into a set of non-empty voxels \(V\), where each voxel \(v_i \in V\) is characterized by a feature vector \(f_i \in \mathbb{R}^C\) encompassing its geometric (e.g., coordinates) and reflectivity properties. The radial masking function, denoted as \(M: V \rightarrow \{0, 1\}\), is a two-stage process that operates on the cylindrical coordinates of each voxel \(v_i\), represented as \((r_i, \theta_i, z_i)\), where \(r_i\) is the radial distance, \(\theta_i\) is the azimuth angle, and \(z_i\) is the height.


\textit{Stage 1: Angular Group Selection:} Voxels are grouped based on their azimuth angle \(\theta_i\) into \(N_g\) angular groups, where each group spans an angular range of \(\Delta \theta = \frac{2\pi}{N_g}\). A subset of these groups is randomly selected with a selection probability \(p_g = 1 - m\), where \(m\) is the desired masking ratio. Let  \(G_s \subset \{1, 2, ..., N_g\}\) denote the indices of the selected groups.

\textit{Stage 2: Range-Aware Masking within Selected Groups:} Within each selected group \(g_j \in G_s\), voxels are further divided into \(N_d\) distance subgroups based on their radial distance \(r_i\). The distance ranges for these subgroups are defined by thresholds \(r_{t_1}, r_{t_2}, ... , r_{t_{N_d}}\). For each voxel \(v_i\) in a selected group \(g_j\), a masking decision is made based on its distance subgroup \(k(v_i)\) and a range-dependent masking probability \(p_{m_{j,k}}\):

\begin{equation}
M(v_i) =
\begin{cases}
0, & \text{if } g(v_i) \in G_s \text{ and }  \text{Bernoulli}(p_{m_{g(v_i), k(v_i)}}) = 1 \\
1, & \text{otherwise}
\end{cases}
\end{equation}

where \(g(v_i)\) denotes the group index to which voxel \(v_i\) belongs, \(k(v_i)\) denotes the distance subgroup index to which voxel \(v_i\) belongs within its group, \(\text{Bernoulli}(p)\) represents a Bernoulli random variable with success probability \(p\).

Notably, the proposed masking strategy significantly reduces LiDAR's energy consumption in on-field operations. By masking out the sensing of angular blocks from the LiDAR's BEV, as shown in the black regions in \autoref{network}(a), we save energy in all LiDAR operations except motor control. Even when a region is sensed, the pretraining in stage 2 encourages the model to maximize accuracy based only on nearby points. As discussed, in Section 2, accurately sensing objects at a distance $R$ requires the laser power $P_\text{laser}$ to increase as $R^4$, thus the laser wave's energy (and thereby laser power) can be dramatically minimized by only relying on the accuracy of nearby points. Additionally, most modern LiDAR systems offer programmable interfaces that can implement the proposed R-MAE during runtime. For example, the Velodyne VLP-16 provides programmable scan pattern interfaces, while the Ouster SDK includes functions to set horizontal and vertical resolution and field of view. These systems can selectively activate lasers for randomly generated angular regions during inference, generating the masked information using pre-trained models.

\subsection{\textbf{Spatially Sparse Convolutional Encoder}} Our encoder leverages 3D sparse convolutions \cite{spconv2022} to efficiently process the masked LiDAR point cloud data. This approach offers several key advantages over Transformer-based alternatives. Sparse convolutions operate only on non-empty voxels, drastically reducing memory consumption and accelerating computations compared to dense operations. This is crucial for handling large-scale 3D scenes and enabling real-time processing for autonomous systems. Unlike Transformer-based methods that flatten 3D point clouds into 2D pillars \cite{hess2023masked, xu2023mv}, sparse convolutions explicitly operate in 3D space, preserving the inherent geometric structure of the scene. This enables the model to learn more nuanced spatial relationships between objects.

The encoder, denoted as \(E: V_s \times \mathbb{R}^C \rightarrow \mathbb{R}^L\), transforms the input features \(f_i \in \mathbb{R}^C\) of the unmasked voxels \(v_i \in V_s\) into a lower-dimensional latent representation \(z_i \in \mathbb{R}^L\).  This transformation is achieved through a series of sparse convolutional blocks, each incorporating 3D convolution, batch normalization, and ReLU activation. Residual connections are also employed to facilitate the training of deep networks and improve gradient flow. The resulting latent representation \(z_i\) encapsulates the learned geometric and semantic features, which are then passed to the decoder for the reconstruction of the masked regions.

\subsection{\textbf{3D Deconvolutional Decoder}} The decoder, denoted as  \(D: \mathbb{R}^L \rightarrow \mathbb{R}^{|V|}\), reconstructs the 3D scene by predicting the occupancy probability \(\hat{o}_i\) for each voxel \(v_i\), including those masked during encoding. It operates on the latent representation \(z_i \in \mathbb{R}^L\) produced by the encoder, progressively recovering spatial information through a series of 3D transposed convolutions (deconvolutions). Each deconvolution layer is followed by batch normalization and ReLU activation, and they collectively upsample the feature maps, increasing spatial resolution until the original voxel grid is reconstructed.  The final layer outputs the predicted occupancy probability \(\hat{o}_i\) for each voxel, which is then compared to the ground truth occupancy \(o_i\) using a binary cross-entropy loss to guide the learning process. By reconstructing the masked regions, the decoder encourages the encoder to learn a compact representation that captures essential geometric and semantic information, crucial for accurate 3D object detection.

\subsection{\textbf{Occupancy Loss}} We adopt occupancy prediction as a pretext task for large-scale point cloud pre-training, building upon the success of ALSO \cite{boulch2023also} and VoxelNet \cite{zhou2018voxelnet} in 3D reconstruction. Occupancy estimation in our model goes beyond mere surface reconstruction; it aims to capture the essence of objects and their constituent parts. By predicting occupancy within a spherical region around each support point, we encourage the model to learn global features representative of different object categories. This fosters a deeper semantic understanding of the point cloud, aiding downstream classification and detection tasks. Occupancy prediction in this context is framed as a binary classification problem due to the prevalence of empty voxels in outdoor scenes and our deliberate partial sensing to conserve energy. The binary cross-entropy loss with logits (BCEWithLogitsLoss) is used to supervise the reconstruction:

\begin{equation}
\begin{split}
L_{\text{occup}} = -\frac{1}{|B|} \sum_{i \in B} \frac{1}{|Q_s|} \sum_{q \in Q_s} \big[ & o_q^i \log(\sigma(\hat{o}_{q}^{i})) \\
& + (1-o_q^i) \log(1-\sigma(\hat{o}_{q}^{i})) \big]
\end{split}
\end{equation}

where \(\hat{o}_{q}^{i}\) is the estimated probability of query voxel $q$ of the $i$-th training sample while \(o_q^i\) is the corresponding ground truth occupancy (1 for occupied, 0 for empty). \(\sigma\) is the sigmoid function. \(|B|\) corresponds to the batch size and \(|Q_s|\) is the number of query voxels in the sphere centered on \(S\). This loss encourages the model to predict accurate occupancy probabilities.

Using the above pretraining strategy, R-MAE strives to maintain spatial continuity in LiDAR scans, while sparse convolutions capture the scene's inherent geometric structure. Additionally, masking entire angular sectors fosters the learning of features robust to yaw rotations, enhancing generalization to unseen viewpoints. These advantages are grounded in the information bottleneck principle \cite{tishby2000information}, which states that the masking process forces the model to extract the most relevant information for reconstruction: $I(X; Z) \leq I(X; \hat{X})$ where \(X\) is the input, \(Z\) is the latent representation, and \(\hat{X}\) is the reconstruction. This combination of factors empowers R-MAE to learn powerful representations for 3D reconstruction and significantly boost prediction accuracy.

\section{Experiments}
\subsection{\textbf{Datasets}} We utilize three major robotics and autonomous driving datasets in our experiments: KITTI 3D \cite{geiger2012we}, Waymo \cite{sun2020scalability}, and nuScenes \cite{caesar2020nuscenes}. KITTI features 7,481 training and 7,518 testing samples with 3D bounding box annotations limited to the front camera's Field of View (FoV), evaluated using mean average precision (mAP) across Easy, Moderate, and Hard difficulty levels. The Waymo Open Dataset includes 798 training sequences (158,361 LiDAR scans) and 202 validation sequences (40,077 LiDAR scans). We subsample 20\% (approximately 32,000 frames) for self-supervised pre-training and finetune on both 20\% and 100\% of the data, using mAP and mAP weighted by heading (APH) metrics at two difficulty levels: L1 and L2. The nuScenes dataset provides 28,130 training and 6,019 validation samples, evaluated with the nuScenes Detection Score (NDS) and metrics such as mAP, average translation error (ATE), average scale error (ASE), average orientation error (AOE), average velocity error (AVE), and average attribute error (AAE).

\begin{table*}[htbp]
\centering
\caption{Quantitative analysis of detection accuracy on the Waymo validation set for models trained on 20\% of the Waymo training data.}
\label{waymo-20}
{\scriptsize
\setlength{\tabcolsep}{3pt} 
\begin{tabular}{l|c|c|c|c|c|c|c}
\toprule
\multirow{2}{*}{Method} & \multicolumn{1}{c|}{mAP/mAPH} & \multicolumn{1}{c|}{Vec\_L1} & \multicolumn{1}{c|}{Vec\_L2} & \multicolumn{1}{c|}{Ped\_L1} & \multicolumn{1}{c|}{Ped\_L2} & \multicolumn{1}{c|}{Cyc\_L1} & \multicolumn{1}{c}{Cyc\_L2} \\
 & L2 & AP/APH & AP/APH & AP/APH & AP/APH & AP/APH & AP/APH \\
\midrule
CenterPoint \cite{yin2021center} & 64.51 / 61.92 & 71.33 / 70.76 & 63.16 / 62.65 & 72.09 / 65.49 & 64.27 / 58.23 & 68.68 / 67.39 & 66.11 / 64.87 \\
+ R-MAE (Ours) & \underline{\textbf{67.20$^{\textcolor{teal}{+2.69}}$ / 64.76$^{\textcolor{teal}{+2.84}}$}} & \underline{\textbf{73.38 / 72.85}} & \underline{\textbf{65.28 / 64.79}} & \underline{\textbf{74.84 / 68.68}} & \underline{\textbf{66.90 / 61.24}} & \underline{\textbf{72.05 / 70.84}} & \underline{\textbf{69.43 / 68.26}} \\
\midrule
PV-RCNN \cite{shi2020pv} & 64.84 / 60.86 & 75.41 / 74.74 & 67.44 / 66.80 & 71.98 / 61.24 & 63.70 / 53.95 & 65.88 / 64.25 & 63.39 / 61.82 \\
+ R-MAE (Ours) & \underline{\textbf{68.95$^{\textcolor{teal}{+4.11}}$ / 66.45$^{\textcolor{teal}{+5.59}}$}} & \underline{\textbf{76.72 / 76.22}} & \underline{\textbf{68.38 / 67.92}} & \underline{\textbf{78.19 / 71.74}} & \underline{\textbf{69.63 / 63.68}} & \underline{\textbf{72.44 / 70.32}} & \underline{\textbf{68.84 / 67.76}} \\
\midrule
Voxel-RCNN \cite{deng2021voxel} & 68.57 / 66.18 & 76.13 / 75.66 & 68.18 / 67.74 & 78.20 / 71.98 & 69.29 / 63.59 & 70.75 / 69.68 & 68.25 / 67.21 \\
+ R-MAE (Ours) & \underline{\textbf{69.09$^{\textcolor{teal}{+0.52}}$ / 66.74$^{\textcolor{teal}{+0.56}}$}} & \underline{\textbf{76.35 / 75.88}} & \underline{\textbf{68.21 / 67.79}} & \underline{\textbf{78.60 / 72.56}} & \underline{\textbf{69.93 / 64.35}} & \underline{\textbf{71.74 / 70.65}} & \underline{\textbf{69.13 / 68.08}} \\
\bottomrule
\end{tabular}
}
\end{table*}

\begin{table*}[htbp]
\centering
\caption{Quantitative analysis of detection accuracy on the Waymo validation set with models trained on 100\% of the Waymo training data.}
\label{waymo-100}
{\scriptsize
\setlength{\tabcolsep}{3pt} 
\begin{tabular}{l|c|c|c|c|c|c|c}
\toprule
\multirow{2}{*}{Method} & \multicolumn{1}{c|}{mAP/mAPH} & \multicolumn{1}{c|}{Vec\_L1} & \multicolumn{1}{c|}{Vec\_L2} & \multicolumn{1}{c|}{Ped\_L1} & \multicolumn{1}{c|}{Ped\_L2} & \multicolumn{1}{c|}{Cyc\_L1} & \multicolumn{1}{c}{Cyc\_L2} \\
 & L2 & AP/APH & AP/APH & AP/APH & AP/APH & AP/APH & AP/APH \\
\midrule
PV-RCNN \cite{shi2020pv} & 69.60 / 67.13 & 78.00 / 77.50 & 69.43 / 68.98 & 79.21 / 73.03 & 70.42 / 64.72 & 71.46 / 70.27 & 68.95 / 67.79 \\
+ R-MAE (Ours) & \underline{\textbf{70.09$^{\textcolor{teal}{+0.49}}$ / 67.75$^{\textcolor{teal}{+0.62}}$}} & \underline{\textbf{78.10 / 77.65}} & \underline{\textbf{69.69 / 69.25}} & \underline{\textbf{79.61 / 73.69}} & \underline{\textbf{71.26 / 65.72}} & \underline{\textbf{71.94 / 70.87}} & \underline{\textbf{69.32 / 68.28}} \\
\bottomrule
\end{tabular}
}
\end{table*}

\begin{table*}[htbp]
\centering
\caption{Quantitative analysis of detection accuracy on the Waymo validation set with models pre-trained on Waymo training data.}
\label{waymo-comparison}
{\scriptsize
\setlength{\tabcolsep}{3pt} 
\begin{tabular}{l|c|c|c|c|c|c|c}
\toprule
\multirow{2}{*}{Method} & \multicolumn{1}{c|}{mAP/mAPH} & \multicolumn{1}{c|}{Vec\_L1} & \multicolumn{1}{c|}{Vec\_L2} & \multicolumn{1}{c|}{Ped\_L1} & \multicolumn{1}{c|}{Ped\_L2} & \multicolumn{1}{c|}{Cyc\_L1} & \multicolumn{1}{c}{Cyc\_L2} \\
 & L2 & AP/APH & AP/APH & AP/APH & AP/APH & AP/APH & AP/APH \\
\midrule
CenterPoint \cite{yin2021center} & 64.51 / 61.92 & 71.33 / 70.76 & 63.16 / 62.65 & 72.09 / 65.49 & 64.27 / 58.23 & 68.68 / 67.39 & 66.11 / 64.87 \\
+ GCC-3D \cite{liang2021exploring} & 65.29$^{\textcolor{teal}{+0.78}}$ / 62.79$^{\textcolor{teal}{+0.87}}$ & - & 63.97 / 63.47 & - & 64.23 / 58.47 & - & 67.68 / 66.44 \\
+ PropCount \cite{yin2022proposalcontrast} & 66.42$^{\textcolor{teal}{+1.91}}$ / 63.85$^{\textcolor{teal}{+1.93}}$ & - & 64.94 / 64.42 & - & 66.13 / 60.11 & - & 68.19 / 67.01 \\
+ Occ-MAE \cite{min2023occupancy} & 65.86$^{\textcolor{teal}{+1.35}}$ / 63.23$^{\textcolor{teal}{+1.31}}$ & 71.89 / 71.33 & 64.05 / 63.53 & 73.85 / 67.12 & 65.78 / 59.62 & 70.29 / 69.03 & 67.76 / 66.53 \\
+ R-MAE (Ours) & \underline{\textbf{67.20$^{\textcolor{teal}{+2.69}}$ / 64.76$^{\textcolor{teal}{+2.84}}$}} & \underline{\textbf{73.38 / 72.85}} & \underline{\textbf{65.28 / 64.79}} & \underline{\textbf{74.84 / 68.68}} & \underline{\textbf{66.90 / 61.24}} & \underline{\textbf{72.05 / 70.84}} & \underline{\textbf{69.43 / 68.26}} \\
\midrule
PV-RCNN \cite{shi2020pv} & 64.84 / 60.86 & 75.41 / 74.74 & 67.44 / 66.80 & 71.98 / 61.24 & 63.70 / 53.95 & 65.88 / 64.25 & 63.39 / 61.82 \\
+ GCC-3D \cite{min2023occupancy} & 61.30$^{\textcolor{teal}{-3.54}}$ / 58.18$^{\textcolor{teal}{-2.68}}$ & - & 65.65 / 65.10 & - & 55.54 / 48.02 & - & 62.72 / 61.43 \\
+ PropCount \cite{yin2022proposalcontrast} & 62.62$^{\textcolor{teal}{-2.22}}$ / 59.28$^{\textcolor{teal}{-1.58}}$ & - & 66.04 / 65.47 & - & 57.58 / 49.51 & - & 64.23 / 62.86 \\
+ MAELi \cite{krispel2024maeli} & - / 62.14$^{\textcolor{teal}{+1.28}}$ & - & - / 67.34 & - & - / 56.32 & - & - / 62.76 \\
+ Occ-MAE \cite{min2023occupancy} & 66.17$^{\textcolor{teal}{+1.33}}$ / 61.98$^{\textcolor{teal}{+1.12}}$ & 75.94 / 75.28 & 67.94 / 67.34 & 74.02 / 63.48 & 64.94 / 55.57 & 67.21 / 66.49 & 65.62 / 63.02 \\
+ R-MAE (Ours) & \underline{\textbf{68.95$^{\textcolor{teal}{+4.11}}$ / 66.45$^{\textcolor{teal}{+5.59}}$}} & \underline{\textbf{76.72 / 76.22}} & \underline{\textbf{68.38 / 67.92}} & \underline{\textbf{78.19 / 71.74}} & \underline{\textbf{69.63 / 63.68}} & \underline{\textbf{72.44 / 70.32}} & \underline{\textbf{68.84 / 67.76}} \\
\bottomrule
\end{tabular}
}
\end{table*}

\subsection{\textbf{Implementation Details}} We evaluate our approach on two key robotics and autonomous driving tasks, object detection and domain adaption using OpenPCDet \cite{openpcdet2020} framework (version 0.6.0). Initially, the R-MAE model undergoes pre-training on the training sets of KITTI, Waymo, and nuScenes datasets without any label exposure. Subsequent fine-tuning on labeled data refines these models further. The process utilizes a pre-trained 3D encoder to start and adjust the backbone networks for these tasks during fine-tuning. The training follows the parameter settings of the original models aligned with OpenPCDet. R-MAE's pre-training involves different masking ratios and angular ranges for voxel processing, aiming to test the effectiveness of the features learned under various configurations during a 30-epoch phase. 

\subsection{\textbf{3D Object Detection}}
We assessed R-MAE for object detection using the Waymo validation set. Pre-training was conducted with 20\% of training data, followed by fine-tuning with various detection heads on both 20\% and 100\% of training dataset. Results are detailed in \autoref{waymo-20} and \autoref{waymo-100} respectively. Using 20\% of the training data for fine-tuning, our pre-trained model achieved mAP improvements of 0.52\% to 4.11\% and mAPH improvements of 0.56\% to 5.59\% over models trained from scratch averaged across all object categories at level-2 difficulty. Fine-tuning with 100\% of the training data yielded gains of 0.49\% to 0.62\% using the same pre-trained weights. These results demonstrate the effectiveness of our pre-training approach in enhancing downstream tasks with limited pre-training data. Additionally, as shown in \autoref{waymo-comparison}, R-MAE's performance surpassed other pre-training methods, particularly in detecting small objects. This enhanced capability is due to the novel radial masking strategy and occupancy reconstruction technique used during pre-training, which improves detection performance by filling in gaps in the representation of smaller objects.

We also assessed R-MAE's performance on the nuScenes dataset, with the outcomes and improvements over baseline training detailed in \autoref{nuescene}. Our model, pre-trained on the nuScenes LiDAR data, underwent fine-tuning in two distinct experiment types. The first set focused on LiDAR-only models, specifically CenterPoint \cite{yin2021center} and Transfusion \cite{bai2022transfusion}, achieving improvements of 2.31\% to 3.17\% in NDS and mAP metrics respectively with our pre-trained weights. Additionally, we explored a multi-modal approach using BEVFusion \cite{liu2023bevfusion}, which combines LiDAR and camera data, though we pre-trained only the LiDAR component. This multi-modal model saw modest gains of 0.49\% and 0.21\%. These results underscore the advantages of applying our pre-trained weights, demonstrating notable benefits even when utilized to prime just one branch of a multi-modal framework. We also assessed R-MAE's performance against other pre-training methods fine-tuned with CenterPoint. Results in \autoref{nuescene-2} demonstrate that our method outperforms the alternatives.

\begin{table*}[htbp]
\centering
\caption{Quantitative performance achieved by different methods on the nuscenes val set.}
\label{nuescene}
{\scriptsize
\begin{tabular}{l|c|c|c|c|c|c|c|c}
\toprule
\textbf{Method} & \textbf{Data Modality} & \textbf{mAP↑} & \textbf{NDS↑} & \textbf{mATE↓} & \textbf{mASE↓} & \textbf{mAOE↓} & \textbf{mAVE↓} & \textbf{mAAE↓} \\
\midrule
CenterPoint \cite{yin2021center} & LiDAR & 56.03 & 64.54 & 30.11 & 25.55 & 38.28 & 21.94 & 18.87 \\
+ R-MAE (Ours) & LiDAR & \underline{\textbf{59.20}$^{\textcolor{teal}{+3.17}}$} &  \underline{\textbf{66.85}$^{\textcolor{teal}{+2.31}}$} & 29.73 & 25.71 & 34.16 & 20.02 & 17.91 \\
\midrule
Transfusion \cite{bai2022transfusion} & LiDAR & 64.58 & 69.43 & 27.96 & 25.37 & 29.35 & 27.31 & 18.55 \\
+ R-MAE (Ours) & LiDAR & \underline{\textbf{65.01}$^{\textcolor{teal}{+0.43}}$} &  \underline{\textbf{70.17}$^{\textcolor{teal}{+0.74}}$} & 28.19 & 25.20 & 26.92 & 24.27 & 18.71 \\
\midrule
BEVFusion \cite{liu2023bevfusion} & LiDAR + Camera & 65.91 & 70.20 & 28.26 & 25.43 & 28.88 & 26.80 & 18.67 \\
+ R-MAE (Ours) & LiDAR + Camera & \underline{\textbf{66.40}$^{\textcolor{teal}{+0.49}}$} &  \underline{\textbf{70.41}$^{\textcolor{teal}{+0.21}}$} & 28.31 & 25.54 & 29.57 & 25.87 & 18.60 \\
\bottomrule
\end{tabular}
}
\end{table*}

\begin{table*}[htbp]
\centering
\caption{Quantitative detection performance achieved by different pre-trained methods on the nuScenes validation set.}
\label{nuescene-2}
{\scriptsize
\setlength{\tabcolsep}{3pt} 
\begin{tabular}{l|c|c|c|c|c|c|c|c}
\toprule
\textbf{Method} & \textbf{Data Modality} & \textbf{mAP↑} & \textbf{NDS↑} & \textbf{mATE↓} & \textbf{mASE↓} & \textbf{mAOE↓} & \textbf{mAVE↓} & \textbf{mAAE↓} \\
\midrule
CenterPoint \cite{yin2021center} & LiDAR & 56.03 & 64.54 & 30.11 & 25.55 & 38.28 & 21.94 & 18.87 \\
+ GCC-3D \cite{liang2021exploring} & LiDAR & 57.3$^{\textcolor{teal}{+1.3}}$ & 65.0$^{\textcolor{teal}{+0.5}}$ & - & - & - & - & - \\
+ BEV-MAE \cite{lin2022bev} & LiDAR & 57.2$^{\textcolor{teal}{+1.2}}$ & 65.1$^{\textcolor{teal}{+0.6}}$ & - & - & - & - & - \\
+ Occupancy-MAE \cite{min2023occupancy} & LiDAR & 57.8$^{\textcolor{teal}{+1.8}}$ & 65.9$^{\textcolor{teal}{+1.4}}$ & 28.8 & 24.4 & 37.5 & 20.6 & 18.0 \\
+ R-MAE (Ours) & LiDAR & \underline{\textbf{59.20}$^{\textcolor{teal}{+3.17}}$} & \underline{\textbf{66.85}$^{\textcolor{teal}{+2.31}}$} & 29.73 & 25.71 & 34.16 & 20.02 & 17.91 \\
\bottomrule
\end{tabular}
}
\end{table*}

Lastly, we present the performance of our R-MAE method on the KITTI validation set in \autoref{kitti-val-40}. Compared to training state-of-the-art models like SECOND \cite{yan2018second} and PVRCNN \cite{shi2020pv} from scratch, R-MAE shows a performance improvement of 0.1\% to 4.3\%, particularly with smaller objects such as cyclists and pedestrians. In addition, our comparisons with other pre-trained models reveal very close average precision (AP) for car detection and improved performance by 1.2\% to 1.6\% for pedestrian and cyclist categories. Although ALSO \cite{boulch2023also} uses a similar pretext task for pre-training focused on occupancy prediction, R-MAE enhances this approach by leveraging masked point clouds for scene construction, using a 3D MAE backbone. This method allows for deeper semantic understanding, improving detection accuracy. Furthermore, our model advances beyond Occupancy-MAE \cite{min2023occupancy} by employing a radial masking algorithm rather than random patch masking, making the MAE backbone more suited for generative tasks and efficient sensing operation on edge devices. Please note that in all tables, numbers in bold represent the results from our R-MAE model, while underlined numbers signify the result of the best performing model. 

\begin{table*}[htbp]
\centering
\caption{Performance comparison on the KITTI $val$ split evaluated by the AP with 40 recall positions at moderate difficulty level. \textsuperscript{\dag}: reproduced by us.}
\label{kitti-val-40}
{\scriptsize
\setlength{\tabcolsep}{2pt} 
\begin{tabular}{l|c|c|c||c|c|c|c}
\toprule
\textbf{Model} & \textbf{Car} & \textbf{Pedestrian} & \textbf{Cyclist} & \textbf{Model} & \textbf{Car} & \textbf{Pedestrian} & \textbf{Cyclist} \\
\midrule
SECOND\textsuperscript{\dag} \cite{yan2018second} & 79.08 & 44.52 & 64.49 & PV-RCNN \cite{shi2020pv} & 82.28 & 51.51 & 69.45 \\
+ Occ.-MAE\textsuperscript{\dag} \cite{min2023occupancy} & \underline{79.12$^{\textcolor{teal}{+0.04}}$} & {45.35$^{\textcolor{teal}{+0.83}}$} & {63.27$^{\textcolor{teal}{-1.22}}$} & + Occ.-MAE\textsuperscript{\dag} \cite{min2023occupancy} & {82.43$^{\textcolor{teal}{+0.15}}$} & {48.13$^{\textcolor{teal}{-3.38}}$} & {71.51$^{\textcolor{teal}{+2.06}}$} \\
+ ALSO\textsuperscript{\dag} \cite{boulch2023also} & {78.98$^{\textcolor{teal}{-0.10}}$} & {45.33$^{\textcolor{teal}{+0.81}}$} & {66.53$^{\textcolor{teal}{+2.04}}$} & + ALSO\textsuperscript{\dag} \cite{boulch2023also} & {82.52$^{\textcolor{teal}{+0.24}}$} & \underline{{52.63$^{\textcolor{teal}{+1.12}}$}} & {70.20$^{\textcolor{teal}{+0.75}}$} \\
+ R-MAE (Ours) & \textbf{79.10$^{\textcolor{teal}{+0.02}}$} & \underline{\textbf{46.93$^{\textcolor{teal}{+2.41}}$}} & \underline{\textbf{67.75$^{\textcolor{teal}{+3.26}}$}} & + R-MAE (Ours) & \underline{\textbf{82.82$^{\textcolor{teal}{+0.54}}$}} & \textbf{51.61$^{\textcolor{teal}{+0.10}}$} & \underline{\textbf{73.82$^{\textcolor{teal}{+4.37}}$}} \\
\bottomrule
\end{tabular}
}
\label{tab:performance_comparison}
\end{table*}

\begin{table*}
\centering
\caption{Quantitative results of domain transfer task. We pre-train R-MAE with Waymo and nuScenes training split and fine-tune with the KITTI training split. Evaluation results are presented at moderate level and 40 recall positions.}
\label{domain}
\scriptsize{
\begin{tabular}{l|c|c|c|c}
\toprule
\textbf{Task} & \textbf{method} & \textbf{Car} & \textbf{Pedestrian} & \textbf{Cyclist} \\
\midrule
 & SECOND \cite{yan2018second} & 79.08 & 44.52 & 64.49 \\
Waymo $\rightarrow$ KITTI & + R-MAE (Ours) & \textbf{79.30$^{\textcolor{teal}{+0.22}}$} & \textbf{48.61$^{\textcolor{teal}{+4.09}}$}  & \textbf{66.62$^{\textcolor{teal}{+2.13}}$}  \\
nuScenes $\rightarrow$ KITTI & + R-MAE (Ours) & \textbf{79.32$^{\textcolor{teal}{+0.24}}$} & \textbf{46.05$^{\textcolor{teal}{+1.53}}$} & \textbf{68.27$^{\textcolor{teal}{+3.78}}$} \\ 
\bottomrule
\end{tabular}
}
\label{tab:performance_comparison}
\end{table*}

\subsection{\textbf{Transferring Domain}} To evaluate the transferability of the learned representation, we fine-tuned R-MAE models on the KITTI dataset using SECOND \cite{yan2018second} and PVRCNN \cite{shi2020pv} as detection bases. As shown in Table \autoref{domain}, the performance gains from R-MAE pre-training are evident in the KITTI domain, indicating that the model acquires a robust, generic representation. Despite the KITTI training samples being smaller compared to Waymo and nuScenes, the R-MAE pre-trained models show significant improvements across different classes. However, the relative improvement is smaller when transferring to KITTI, likely due to the domain gap. This suggests that while R-MAE effectively learns generalizable features, variations in data domains can impact performance improvements.


\subsection{\textbf{R-MAE's Parametric Space Exploration}} 
\begin{figure*}[t!]
    \centering
    \includegraphics[width=0.43\textwidth]{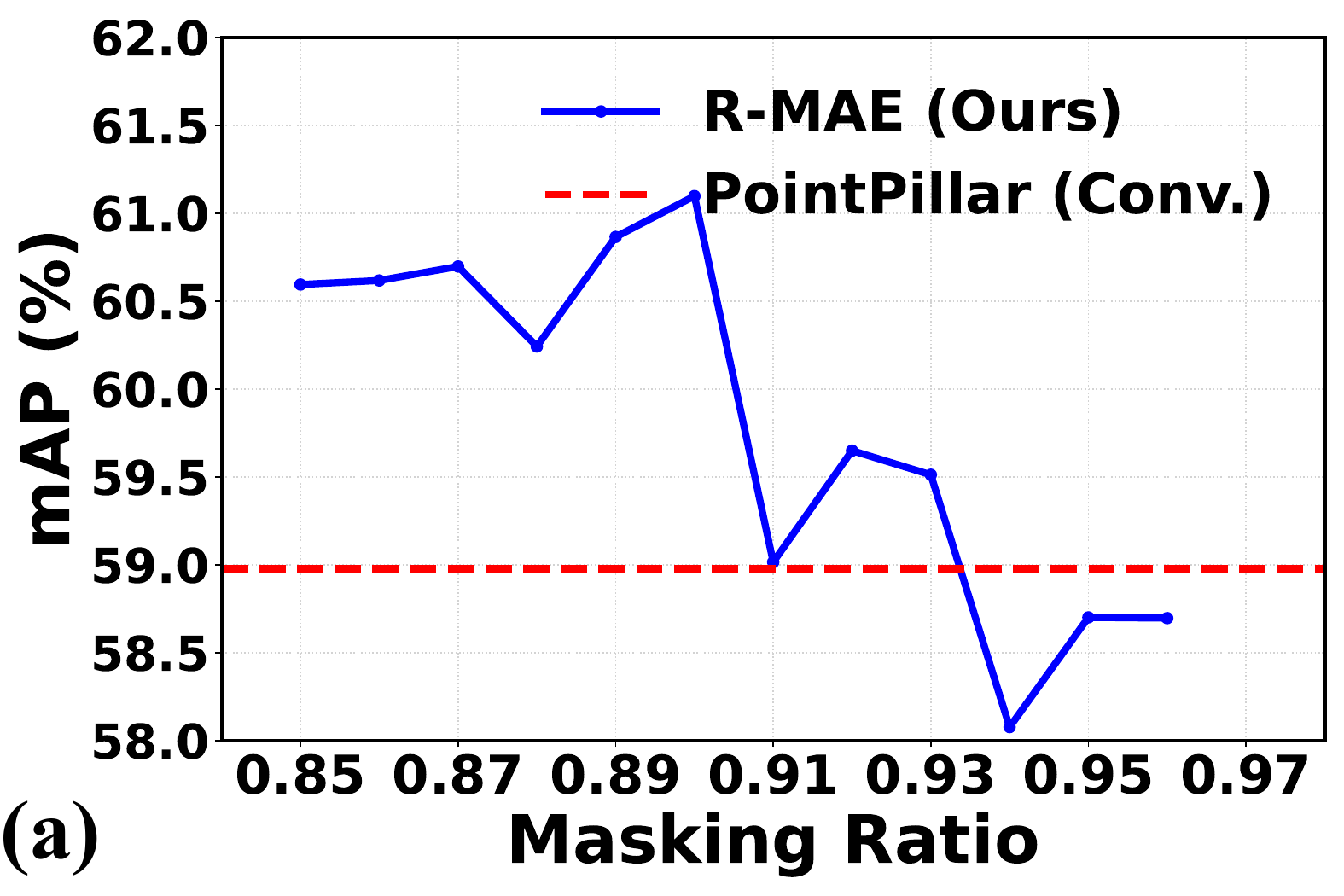}
    \includegraphics[width=0.43\textwidth]{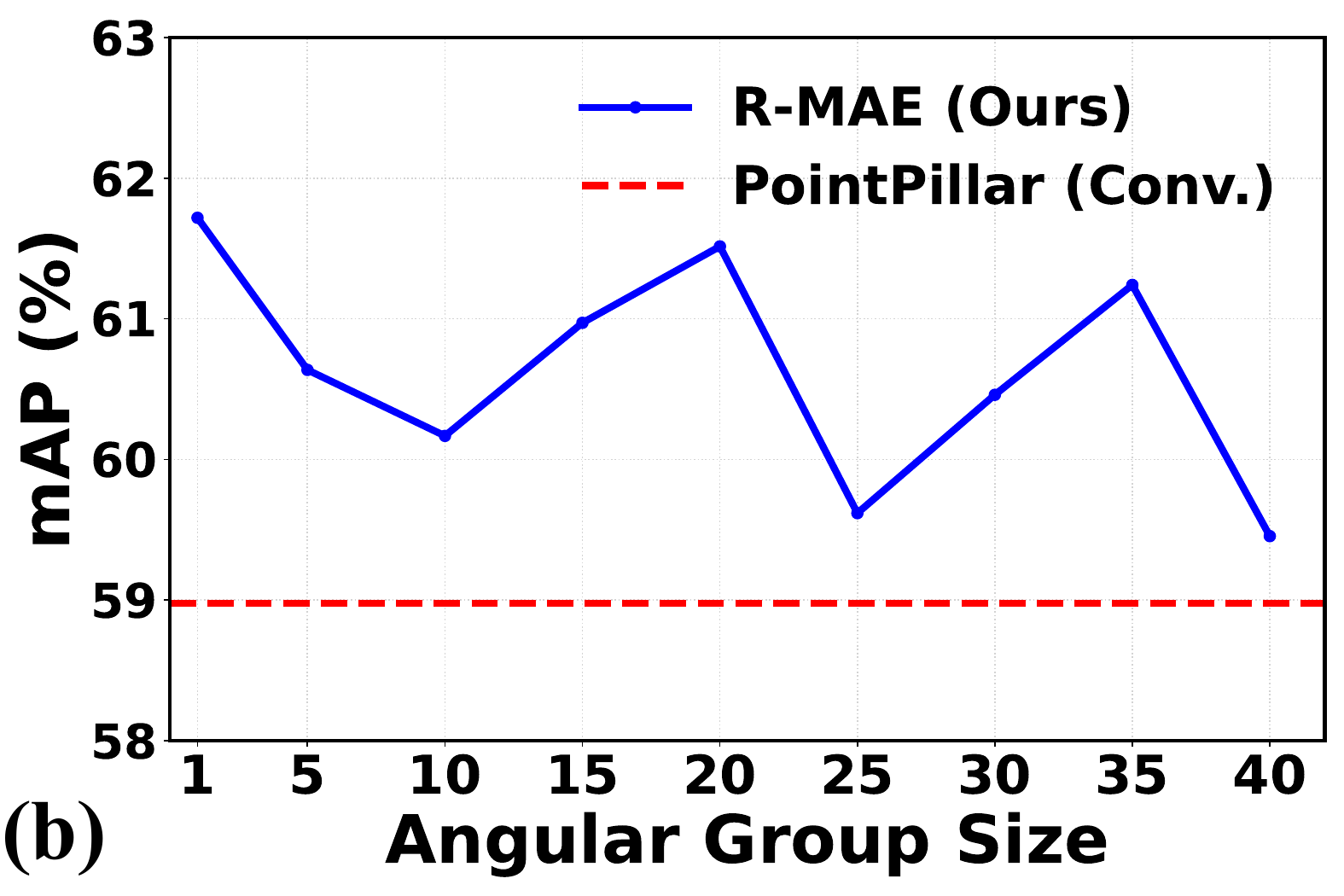}
    \caption{Accuracy under varying pretraining conditions, \textbf{(a)} at varying masking ratios with a fixed angular range of 1 degree, and \textbf{(b)} at different angular ranges with a fixed masking ratio of 0.8. Results are compared against the state-of-the-art (SOTA) PointPillars \cite{lang2019pointpillars} method.}
    \label{ablation}
\end{figure*}

We conducted additional studies to explore the limits of the proposed R-MAE by varying the masking ratio and modulating the size of contiguous angular segments that are not sensed [see these settings in \autoref{ablation}]. All experiments were performed using the pre-trained R-MAE fine-tuned on PointPillar \cite{lang2019pointpillars} detection head. \autoref{ablation}(a) compares the accuracy of the proposed approach against SOTA PointPillar. Notably, the accuracy of our method only begins to gracefully degrade beyond a masking ratio of 0.92, indicating that only 8\% of LiDAR's BEV needs to be sensed, with the rest being generated, thereby enabling ultra-frugal LiDAR operation. In \autoref{ablation}(b), we examine the impact of angular size of voxel grouping before masking. All experiments used an 80\% masking ratio to assess the effect of different angles. On the considered dataset, R-MAE is almost invariant to the angular size of the grouping range; however, angular size dependence may arise for other datasets, likely due to reduced voxel diversity and less varied features after masking at wider angles.

\section{Conclusion}
We demonstrated how R-MAE can trade off sensing energy with training data for low-power robotics and autonomous navigation to operate frugally with sensors. R-MAE-based LiDAR processing \textit{generates} rather than \textit{senses} predictable or inconsequential parts of the environment, enabling ultrafrugal sensing. R-MAE achieves over a 5\% average precision improvement in detection tasks and over a 4\% accuracy improvement in domain transfer on Waymo, nuScenes, and KITTI datasets. It enhances small object detection by up to 4.37\% in AP on the KITTI dataset and surpasses baseline models by up to 5.59\% in mAP/mAPH on the Waymo dataset with 90\% radial masking. Additionally, it achieves up to 3.17\% and 2.31\% improvements in mAP and NDS on nuScenes dataset. 

\noindent\textbf{Acknowledgement:} This work was supported in part by COGNISENSE, one of seven centers in JUMP 2.0, a Semiconductor Research Corporation (SRC) program sponsored by DARPA and NSF Award \#2329096.

\bibliographystyle{IEEEtran}
\bibliography{main}

\end{document}